\documentclass[10pt,twocolumn,letterpaper]{article}

\usepackage{cvpr}              
\usepackage{multirow}
\usepackage[table]{xcolor}
\usepackage{amssymb,pifont}
\providecommand{\XSolidBrush}{\ding{55}}
\usepackage{algorithm}
\usepackage{algorithmic}
\definecolor{cvprblue}{rgb}{0.21,0.49,0.74}
\usepackage{tikz}
\usetikzlibrary{positioning, arrows.meta, shapes.geometric, calc}
\usepackage[pagebackref,breaklinks,colorlinks,allcolors=cvprblue]{hyperref}

\def\paperID{12644} 
\def\confName{CVPR}
\def\confYear{2026}

\title{XSeg: A Large-scale X-ray Contraband Segmentation Benchmark For Real-World Security Screening}

\author{
Hongxia Gao\textsuperscript{1,3,4,*,\dag}, 
Yixin Chen\textsuperscript{2},
Jiali Wen\textsuperscript{2},
Litao Li\textsuperscript{2,*},
Kaijie Zhang\textsuperscript{2},
Qianyun Liu\textsuperscript{2}\\
\textsuperscript{1}Xi'an Jiaotong University,
\textsuperscript{2}South China University of Technology,
\textsuperscript{3}Shenzhen Loop Area Institute, \\
\textsuperscript{4}Pazhou Laboratory\\
\vspace{0.5em}
\href{https://huggingface.co/datasets/xi801/XSeg}{
\includegraphics[height=0.9em]{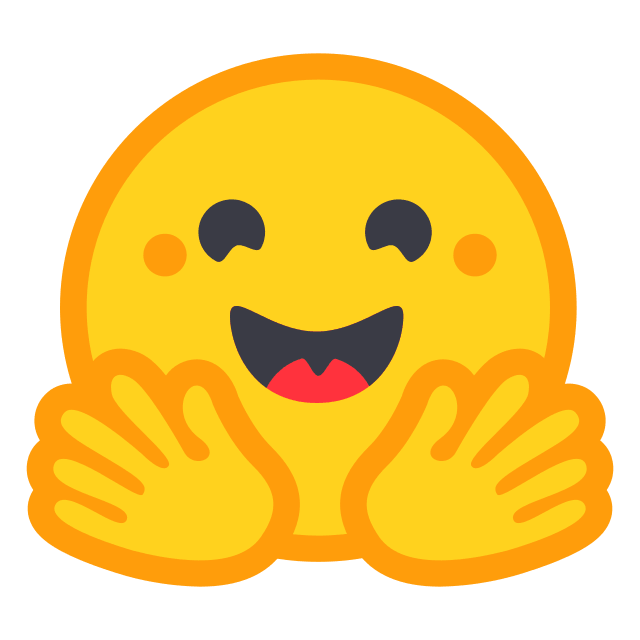}
\hspace{0.4em}XSeg Dataset Page
}
}

\begin{document}
\maketitle
\let\thefootnote\relax\footnotetext{* Equal contribution.}
\let\thefootnote\relax\footnotetext{\dag~Corresponding author. Hongxia Gao. hxgao@xjtu.edu.cn}
\begin{abstract}
X-ray contraband detection is critical for public safety. However, current methods primarily rely on bounding box annotations, which limit model generalization and performance due to the lack of pixel-level supervision and real-world data. To address these limitations, we introduce \textbf{XSeg}. To the best of our knowledge, XSeg is the largest X-ray contraband segmentation dataset to date, including \textbf{98,644} images and \textbf{295,932} instance masks, and contains the latest \textbf{30} common contraband categories. The images are sourced from public datasets and our synthesized data, filtered through a custom data cleaning pipeline to remove low-quality samples. To enable accurate and efficient annotation and reduce manual labeling effort, we propose Adaptive Point SAM (APSAM), a specialized mask annotation model built upon the Segment Anything Model (SAM). We address SAM’s poor cross-domain generalization and limited capability in detecting stacked objects by introducing an Energy-Aware Encoder that enhances the initialization of the mask decoder, significantly improving sensitivity to overlapping items. Additionally, we design an Adaptive Point Generator that allows users to obtain precise mask labels with only a single coarse point prompt. Extensive experiments on XSeg demonstrate the superior performance of APSAM.
\end{abstract}    
\section{Introduction}
\label{sec:intro}

\begin{figure*}[t]
  \centering
  \includegraphics[width=\textwidth]{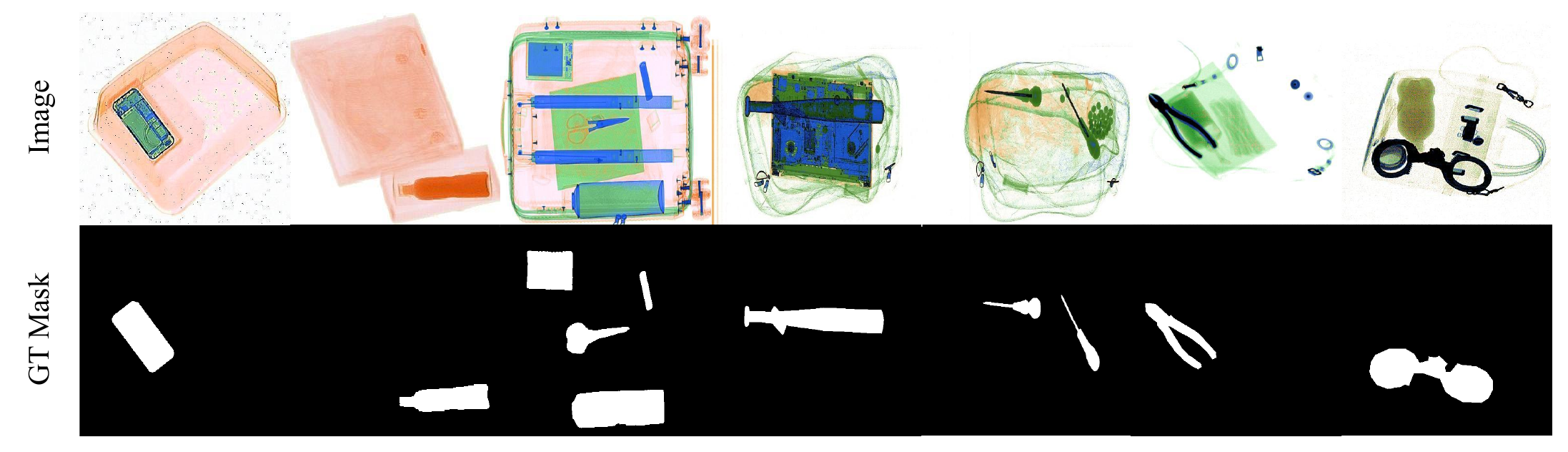}

   \caption{XSeg data examples. The first row shows X-ray images of contraband from various sources, including but not limited to MobilePhone, Liquid, Scissors, Baton, and Handcuffs. Ground-Truth masks are generated by SAM~\cite{sam} and refined by multiple security experts.}
   \label{fig:onecol}
\end{figure*}

\begin{figure}[h]
\centering
\includegraphics[width=\columnwidth]{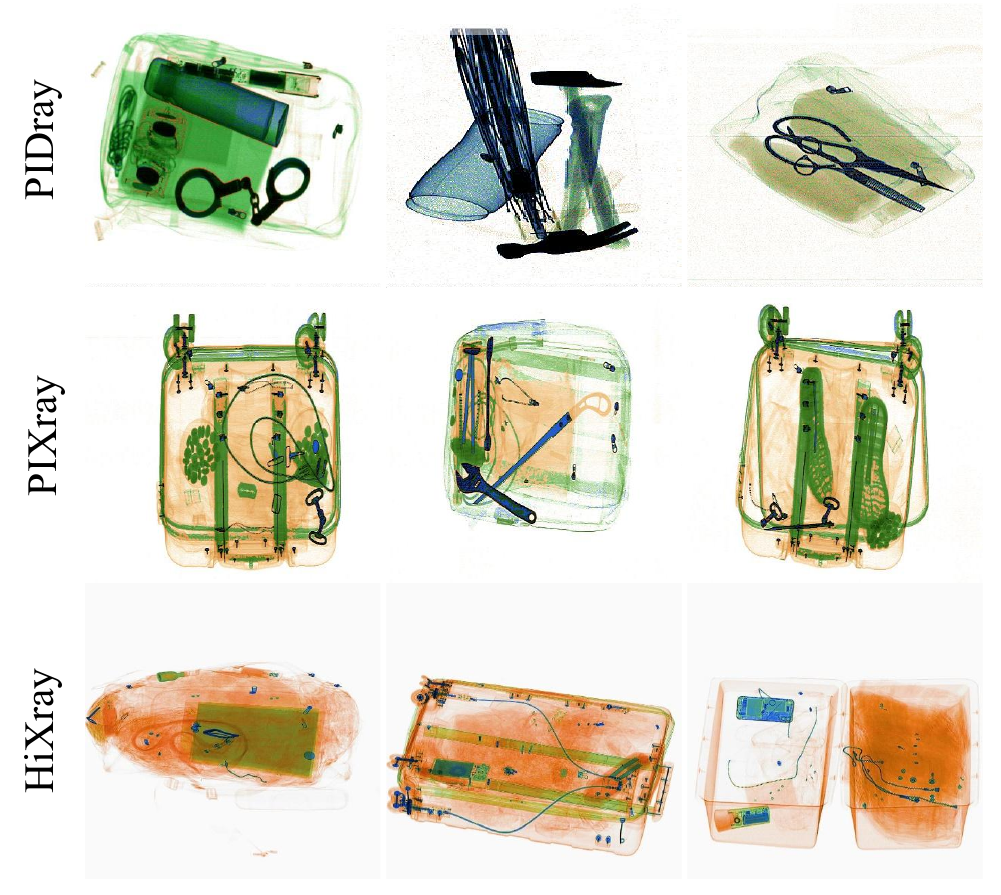}
\caption{Cross-domain chromatic variance phenomenon, the first row is PIDray~\cite{pidray}, the overall color is greenish, the second row is PIXray~\cite{pixray}, the imaging is still greenish, and the third row is HiXray~\cite{hixray}, the color distribution of the luggage is significantly different from PIDray and PIXray.}
\label{fig2}
\end{figure}

X-ray security inspection systems are widely deployed in high-throughput scenarios such as airports and logistics hubs to identify potential threats through baggage screening. While computer-aided detection (CAD) techniques based on deep learning have become operational standards, critical limitations persist in current implementations. The predominant bounding box-based detection paradigm suffers from two fundamental limitations in X-ray imaging environments. First, material superposition creates chromatic distortion at overlapping regions due to density-dependent photon absorption, causing feature ambiguity between occluded objects. Second, coarse bounding annotations introduce extraneous background signals that degrade detection specificity. These characteristics substantially compromise inspection reliability in real-world scenarios with complex item arrangements.Recent efforts to develop segmentation benchmarks~\cite{pixray,pidray,stcray} address annotation granularity but fail to overcome three critical dataset deficiencies: (1) \textbf{Data scarcity and compositional simplicity:} Existing collections contain limited samples (typically less than 50k images) dominated by single-category threats, inadequately representing multi-object occlusion scenarios prevalent in actual baggage; (2) \textbf{Cross-domain chromatic variance}: Current datasets exhibit machine-specific color distributions from single-source acquisition, as shown in Figure~\ref{fig2}. This distribution shift necessitates costly retraining when deploying models across different X-ray scanners; (3) \textbf{Insufficient category coverage:} With expanding security regulations, existing datasets cover less than 21 threat types, failing to address emerging contraband categories like novel electronic concealments. Together, these limitations hinder the ability to generalize the model, and there is an urgent need for solutions to accommodate the above issues.

\begin{table*}[h]
\centering
\caption{Comparison of X-ray Contraband Detection datasets, all datasets in the table are open source and available, \textbf{Real} means the data is from the real world, \textbf{Synthetic} means the data is from synthetic. 'C','D','S','ZS' stand for Classification, Detection, Segmentation, and Zero-Shot tasks, respectively. XSeg has an advantage over the other datasets in terms of categories, the number of images that contain contraband.}
\resizebox{\textwidth}{!}{%
\begin{tabular}{c|ccc|cccc|c|c}
\hline
 &  &  &  & \multicolumn{4}{c|}{Annotations} &  &  \\ \cline{5-8}
\multirow{-2}{*}{Dataset} & \multirow{-2}{*}{Year} & \multirow{-2}{*}{Classes} & \multirow{-2}{*}{Positive} & Classfication & Bbox & Mask & \multicolumn{1}{l|}{Caption} & \multirow{-2}{*}{Type} & \multirow{-2}{*}{Tasks} \\ \hline
GDXray~\cite{gdxray} & 2015 & 3 & 8,150 & \checkmark & \checkmark & \XSolidBrush & \XSolidBrush & Real & C, D \\
SIXray~\cite{sixray} & 2019 & 6 & 8,929 & \checkmark & \checkmark & \XSolidBrush & \XSolidBrush & Real & C, D \\
OPIXray~\cite{opixray} & 2020 & 5 & 8,885 & \checkmark & \checkmark & \XSolidBrush & \XSolidBrush & Synthetic & C, D \\
HiXray~\cite{hixray} & 2021 & 8 & 45,364 & \checkmark & \checkmark & \XSolidBrush & \XSolidBrush & Real & C, D \\
CLCXray~\cite{clcxray} & 2022 & 12 & 9,565 & \checkmark & \checkmark & \XSolidBrush & \XSolidBrush & Real & C, D \\
PIXray~\cite{pixray} & 2022 & 15 & 5,046 & \checkmark & \checkmark & \checkmark & \XSolidBrush & Real & C, D, S \\
PIDray~\cite{pidray} & 2023 & 12 & 47,677 & \checkmark & \checkmark & \checkmark & \XSolidBrush & Real & C, D, S \\
114Xray~\cite{114xray} & 2024 & 12 & 58,000 & \checkmark & \checkmark & \XSolidBrush & \XSolidBrush & Real & C, D, S \\
STCray~\cite{stcray} & 2025 & 21 & 45,693 & \checkmark & \checkmark & \checkmark & \checkmark & Real\&Synthetic & \textbf{C, D, S, ZS} \\ \hline
\rowcolor[HTML]{EFEFEF} 
\textbf{XSeg (Ours)} & 2025 & \textbf{30} & \textbf{98,644} & \checkmark & \checkmark & \checkmark & \XSolidBrush & \textbf{Real\&Synthetic} & C,D,S \\ \hline
\end{tabular}
}
\label{tab1}
\end{table*}

Another critical factor hindering the iteration and expansion of contraband datasets is the inefficiency of current annotation tools. These tools are typically built upon existing detection or segmentation algorithms, yet current segmentation-based X-ray contraband detection methods~\cite{pidray,pixray} remain limited by their reliance on time-consuming pixel-level annotations and insufficient domain adaptability. Although foundation models like Segment Anything (SAM)~\cite{sam} have shown exceptional few-shot segmentation capabilities in natural images via a prompt-driven paradigm, their direct application to X-ray security inspection faces an inherent domain gap: density-dependent absorption imaging in X-ray compromises SAM's natural image priors, while complex object stacking in cluttered luggage disrupts the model's perception of detailed features. Recent domain adaptation strategies, such as lightweight adapters~\cite{lora} and Vision-Language Model fine-tuning~\cite{stcray,mmcl} have partially mitigated these issues through empirical data augmentation or cross-modal alignment. However, they commonly overlook X-ray image physical feature priors, which are crucial for robust feature decomposition in multi-layer occlusion scenarios. This oversight leads to performance degradation when handling scanner-specific chromatic aberrations or novel threat configurations, underscoring the need for domain-specific architectures that intrinsically embed X-ray imaging physics.
Based on these observations, we introduce \textbf{XSeg}, a large-scale real-world X-ray contraband segmentation dataset with category labels, bounding box labels, and mask labels, aiming to reflect the complexity of actual airport and traffic security inspection environments. The dataset integrates three publicly available datasets, including raw image data from 114Xray~\cite{114xray}, PIDray~\cite{pidray}, and PIXray~\cite{pixray}, which collectively capture various types of contraband and complex object stacking scenarios commonly encountered in real-world inspections. These multi-source images exhibit significant variations in imaging conditions and material appearance, thereby enhancing the dataset's generalization capabilities. Considering the limitations of the original datasets, such as inaccurate or rough mask boundaries, incomplete annotations, and coarse-grained category definitions, we implemented a structured data cleaning pipeline. The pipeline includes three automatic filters to remove low-quality and noisy image-mask pairs, followed by a team of professional safety inspection experts who meticulously re-label the entire dataset. The team refined the original category definitions, dividing them into fine-grained classifications, and used an internally developed segmentation model to assist in generating accurate and detailed pixel-level masks. Each annotation undergoes multiple rounds of quality review to ensure precise boundary alignment and semantic consistency. With comprehensive coverage of occlusion levels (from partial to severe occlusion), XSeg provides a realistic benchmark for developing and evaluating segmentation models in security screening applications. To our knowledge, XSeg is currently the largest and most diverse X-ray contraband segmentation dataset, featuring 30 categories and 98,644 images of contraband. 

Leveraging SAM's~\cite{sam}strong few-shot segmentation capabilities, we've developed APSAM, a large-scale segmentation model specifically designed for the X-ray contraband domain, built upon the XSeg foundation. APSAM maintains SAM's core framework, comprising an image encoder, prompt encoder, and mask decoder. We've significantly enhanced it by integrating a lightweight Energy-Aware Encoder (EAE) in parallel with the main image encoder. This EAE utilizes the physical properties of X-ray high and low-energy models to provide crucial initial localization information for occluded objects. Furthermore, recognizing that a single user-provided point prompt often lacks sufficient detail, we designed an Adaptive Point Generator (APG), which adaptively expands a single user point into two, leading to a significant boost in segmentation performance.

In summary, this work primarily consists of the following three contributions:
\begin{itemize}
    \item We introduce XSeg, the largest and most diverse X-ray contraband segmentation benchmark to date, featuring \textbf{98,644} image-mask pairs with detailed pixel-level annotations across \textbf{30} categories and realistic occlusion conditions for real-world security inspection.
    \item  we introduce APSAM, an X-ray contraband segmentation model that enhances performance by integrating an \textbf{Energy-Aware Encoder (EAE)} for occluded object localization and an \textbf{Adaptive Point Generator (APG)} to intelligently expand user prompts.
    \item Extensive experiments demonstrate APSAM's superior performance, outperforming state-of-the-art models on the X-ray contraband segmentation task and achieving leading results on the comprehensive XSeg benchmark, validating its effectiveness and generalization.
\end{itemize}

\section{Related Works}
\label{sec:rw}


\textbf{X-ray Contraband Datasets:} Early X-ray datasets, such as GDXray~\cite{gdxray} and SIXray~\cite{sixray}, were limited by small annotated subsets, coarse bounding box annotations, and narrow category coverage. Recent trends have moved towards fine-grained categorization, broader class coverage, and dense annotations, exemplified by OPIXray~\cite{opixray}, HiXray~\cite{hixray}, and PIXray~\cite{pixray}. While PIDray~\cite{pidray} introduced occlusion-aware stratification, its single-instance focus limits utility in cluttered scenes. The largest current datasets, 114Xray~\cite{114xray} and STCray~\cite{stcray}, offer extensive images and multimodal features, respectively. However, key limitations persist: inadequate mask-level annotation coverage and insufficient data for complex occlusion and multi-instance scenarios, underscoring the critical need for more precise annotations and diverse scene construction to advance real-world X-ray security inspection.
\section{Dataset XSeg}

Current X-ray image datasets suffer from inconsistent color space distributions and limited annotations, which severely constrain the generalization ability of segmentation models. This limitation hinders their effectiveness in real-world contraband detection scenarios, where a wide variety of packages and baggage are encountered. To tackle these challenges, we introduce XSeg, a dataset comprising 98,644 images and 295,932 binary instance masks. To our knowledge, XSeg is currently the largest segmentation dataset in the contraband detection domain. It significantly outperforms existing datasets such as PIDray~\cite{pidray} and PIXray~\cite{pixray} in both image quantity and label quality. Specific information about XSeg and other datasets is presented in Table~\ref{tab1}.

\subsection{Data Curation}

\begin{figure*}[h!]
\centering
\includegraphics[width=\textwidth]{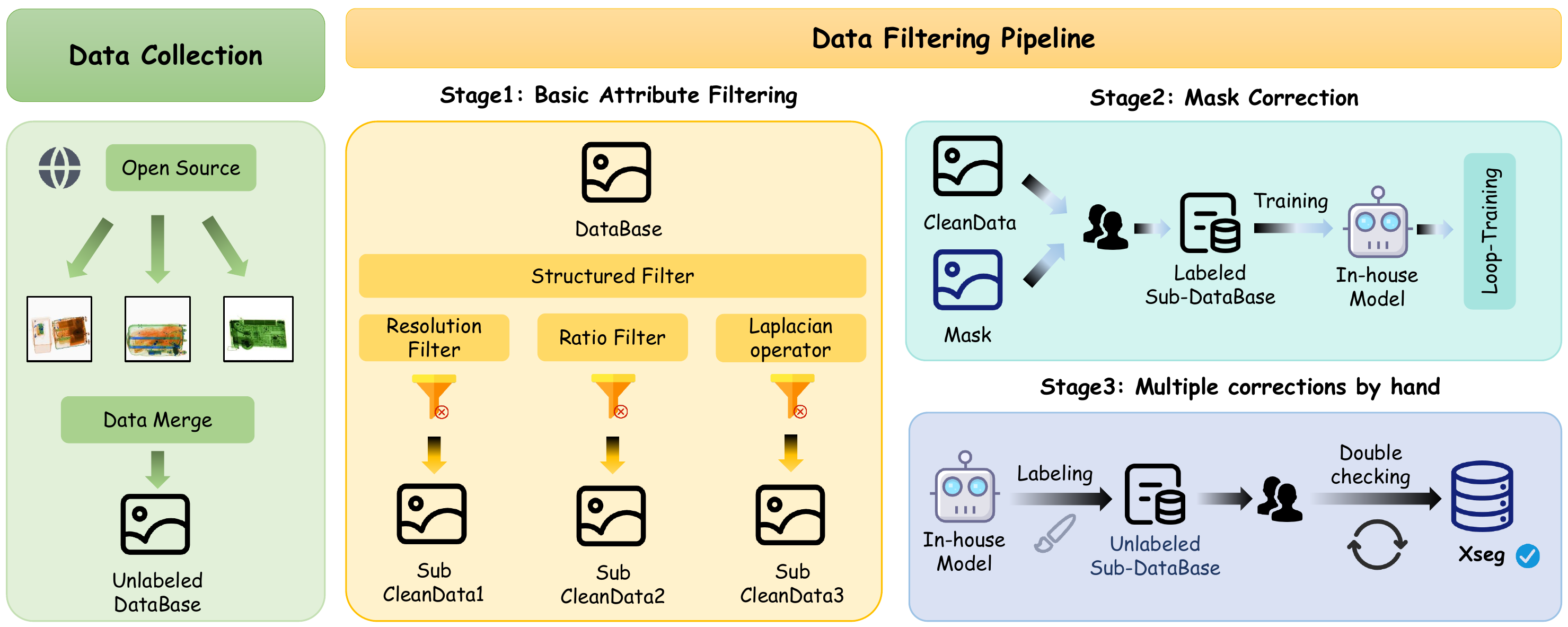} 
\caption{Pipeline of data cleaning. First, images are filtered based on resolution, aspect ratio, and noise levels using Laplacian variance thresholds, effectively removing low-resolution, irregularly proportioned, or excessively noisy samples. The cleaned images are then manually annotated by experts to establish high-precision ground truth masks. These annotations are subsequently used to train an in-house segmentation model, forming an automated labeling system. The model-generated masks undergo multiple rounds of iterative refinement to ensure boundary accuracy, with human verification at each stage. The final output is XSeg.}
\label{figpipe}
\end{figure*}

Our data collection starts with images from three public datasets: 114Xray~\cite{114xray}, PIXray~\cite{pixray}, and PIDray~\cite{pidray}. We further augment them with real-world security screening images captured in operational settings, covering diverse scenarios such as airports, subway stations, logistics parcels, and composite contraband. These sources use different imaging devices and acquisition protocols, leading to inherent shifts in RGB intensity distributions. Rather than a drawback, this domain heterogeneity serves as natural diversity that improves model generalization in end-to-end segmentation. To maintain high fidelity, we apply a systematic curation pipeline that filters out low-quality samples—especially those with very low resolution or severe noise. The complete data collection and cleaning process is shown in Figure~\ref{figpipe}.
Specifically, we initially employ a resolution filter and an aspect ratio filter to discard images with resolutions below 200 pixels or aspect ratios outside the range $[0.2,5]$. Subsequently, we leverage the Laplacian operator to evaluate image sharpness by computing the variance of the Laplacian. An empirical threshold $T$ is set for the variance, and images with variance exceeding $T$ are identified as containing significant noise that may degrade data quality. These noisy images are discarded with a probability of $0.9$. Through this pipeline, the initial dataset of approximately 150k images is refined to a final set of 98,644 high-quality samples, thereby significantly enhancing dataset integrity and training stability.

\subsection{Data Annotation}
Precise boundary delineation is essential for mask labels, surpassing the accuracy of bounding boxes. To achieve pixel-level accurate mask annotations, we assembled a team of experts in security inspection and X-ray imaging and developed an internal annotation platform. After a year of iterative optimization and quality control, we constructed the high-quality XSeg dataset.

For the PIDray~\cite{pidray} and PIXray~\cite{pixray}, which provide instance masks, we used MobileSAM~\cite{mobile_sam} and manual validation to refine the initially rough mask boundaries. The remaining unlabeled images were carefully annotated using SAM~\cite{sam} and manual expertise. To improve the annotation efficiency, we trained a SAM on the binary mask of the initial dataset via an adapter. The model was integrated into the visual annotation platform to generate the initial mask~\cite{114xray} for the 114Xray data. As shown in Figure~\ref{fig2}, despite our improvements, there is still a color gap between the corrected PIDray and PIXray images and the 114Xray data. Therefore, we also manually reviewed and optimized the SAM-generated masks. Specifically, we initially corrected 10,000 image-mask pairs. The corrected data were then re-added to the training set to iteratively improve model performance. This closed-loop annotation strategy was repeated five times to ensure high accuracy of the annotations in XSeg.

To maintain consistency with the original public datasets, we preserved their training/validation/test set splits and utilized these splits in our experiments.


\subsection{Data Split}
To maintain domain-specific characteristics across diverse acquisition protocols, we adopted a source-aware partitioning strategy. For public datasets (114Xray~\cite{114xray}, PIXray~\cite{pixray}, PIDray~\cite{pidray}), we preserved their original validation/test splits after quality filtering. For operational images from airports, subway stations, and logistics centers, we applied an 8:1:1 split. This approach maintains domain integrity while enabling meaningful cross-domain generalization analysis. The final dataset split is detailed in Table~\ref{split}.

\begin{table}[h]
\centering
\caption{XSeg dataset split.}
\label{tab:dataset_split}
\begin{tabular}{ccccc}
\hline
 & Training & Validation & Test & Total \\ \hline
Xseg & 68518 & 23735 & 10019 &98644 \\ \hline
\end{tabular}
\label{split}
\end{table}

\subsection{Additional Version}
In addition to binary instance masks, we introduce XSeg-semantic, a fine-grained semantic annotation version of the dataset. Specifically, we leverage expert knowledge to define detailed category subdivisions and use feature responses from a trained classifier for clustering analysis to validate and refine these classifications. Through this iterative process, we establish 30 semantic classes in XSeg, such as splitting Scissors into MetalhandleScissors and PlastichandleScissors based on handle material (See Appendix), enabling more precise contraband detection and analysis.

\section{Adaptive Point SAM}
\label{sec:APSAM}
\subsection{Segment Anything Model}

The Segment Anything Model (SAM)~\cite{sam} performs few-shot segmentation via an image encoder $E_{\text{img}}$, a prompt encoder $E_{\text{prompt}}$, and a mask decoder $D_M$. Given an image $\mathbf{X}$ and prompt $\mathcal{P}$, SAM computes image embeddings $\mathbf{I}=E_{\text{img}}(\mathbf{X})$ and prompt embeddings $\mathbf{P}=E_{\text{prompt}}(\mathcal{P})$. The decoder iteratively refines output tokens $\mathbf{H}$ through cross-attention with $\mathbf{I}$ over $L$ layers:
\begin{equation}
    \mathbf{H}^{(l)} = D_M\big(\mathbf{H}^{(l-1)}, \mathbf{I}\big), \quad l = 1, \dots, L,
\end{equation}
where $\mathbf{H}^{(0)}=[\mathbf{O}_{\text{iou}}; \mathbf{O}_{\text{mask}}; \mathbf{P}]$. Final masks and IoU scores are predicted by linear heads:
\begin{align}
    s_{\text{iou}} &= \sigma(\mathbf{W}_i \mathbf{O}_{\text{iou}}), \\
    \mathbf{M} &= \mathrm{Sigmoid}(\mathbf{W}_m \mathbf{O}_{\text{mask}}).
\end{align}
The mask with the highest IoU score is selected as output: $\hat{\mathbf{M}}=\mathbf{M}[\arg\max s_{\text{iou}}]$.

\subsection{Framework of APSAM}

\begin{figure*}[h]
\centering
\includegraphics[width=0.8\textwidth]{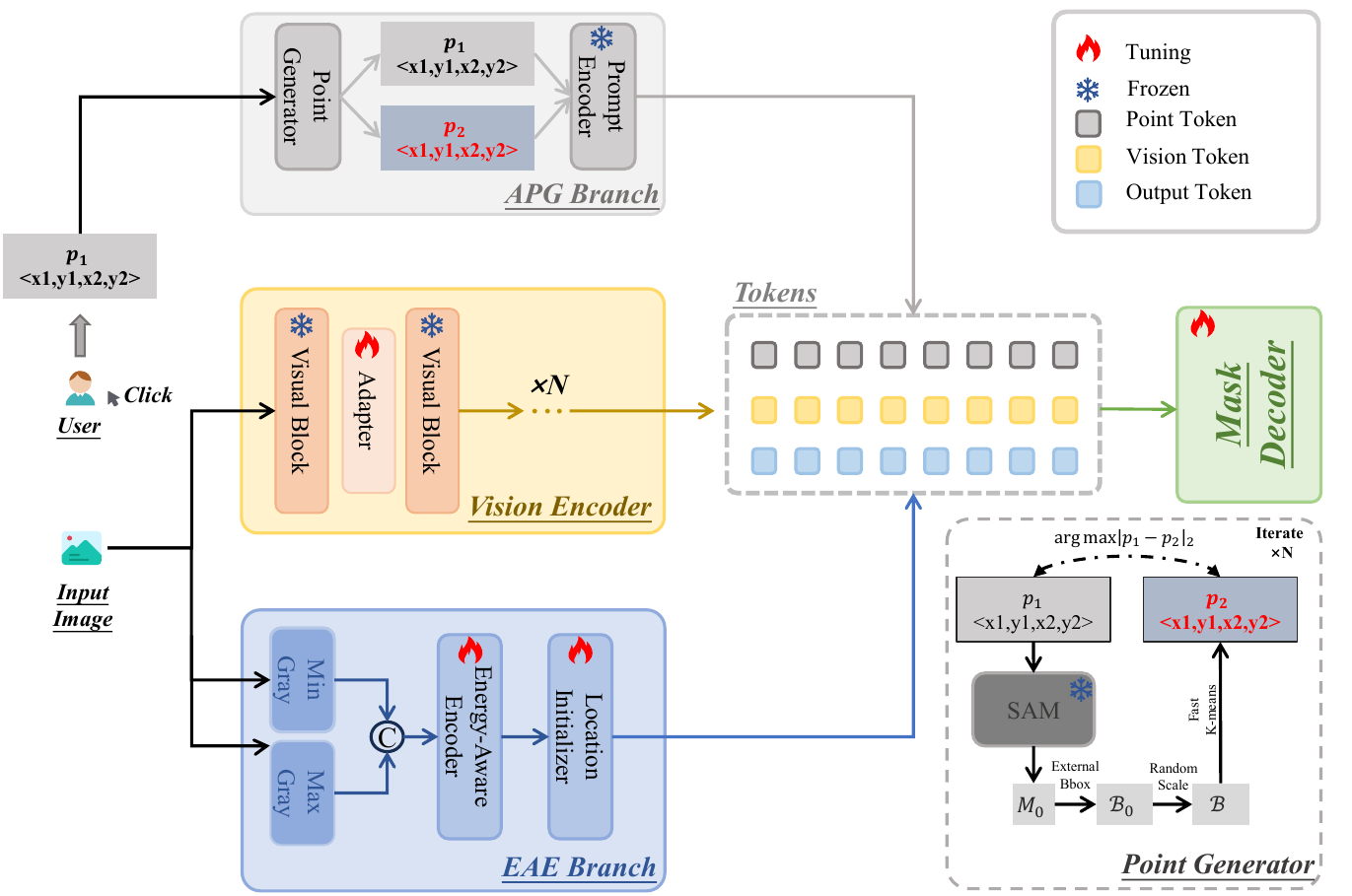}
\caption{Framework of APSAM. APSAM initializes output tokens by concatenating maximum and minimum grayscale image representations. These are then processed by an Energy-Aware Encoder (EAE) and an MLP based Location Initializer. For the visual encoder backbone, we fine-tune with an identical adapter, keeping other parameters frozen. Point prompts are adaptively generated by our APG, and the mask decoder produces the final segmentation.}
\label{apsampipe}
\end{figure*}

This chapter presents APSAM, an optimized segmentation model built on the SAM framework for contraband segmentation. The architecture of APSAM is shown in Figure~\ref{apsampipe}. We've enhanced it with two key components. First, an \textbf{Energy-Aware Encoder} uses expert prior understanding of X-ray images, initializing the Mask Decoder's output token with contraband-aware regions from high and low-energy images. This provides an initial localization of concealed contraband. Second, acknowledging the critical role of prompts, our \textbf{Adaptive Point Generator} creates an additional point prompt for the user-provided point, significantly improving segmentation performance.

\subsection{Energy-Aware Encoder}

X-ray images are inherently stored as atomic number maps. Dual-energy X-ray scanning of contraband yields high- and low-energy projections that capture global material composition and fine structural details respectively. Conventional pseudo-coloring of these grayscale channels into RGB introduces non-linear distortion and channel redundancy that degrade subtle contraband features.

\begin{figure}[t]
\centering
\includegraphics[width=0.6\linewidth]{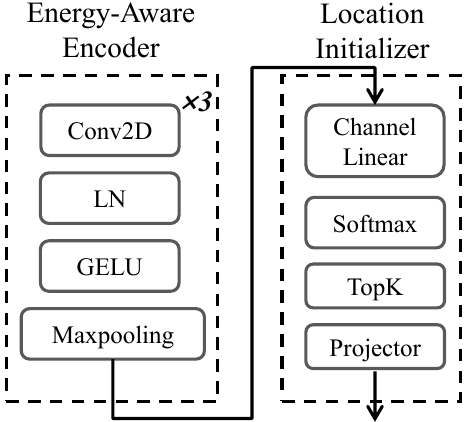}
\caption{Energy-Aware Encoder and Location Initializer. The Energy-Aware Encoder comprises three convolutional layers featuring GELU activation and max-pooling. The Location Initializer primarily relies on linear layers and softmax for channel selection, ultimately employing Top-$k$ filtering to identify the optimal token.}
\label{eae}
\end{figure}

To preserve intrinsic energy-specific information, we propose an Energy-Aware Encoder (EAE). Given an input dual-energy image $I \in \mathbb{R}^{H \times W \times C}$, we derive its high- and low-energy components as
\begin{equation}
    I_H = \max_{c} I(\cdot, \cdot, c), \quad I_L = \min_{c} I(\cdot, \cdot, c),
\end{equation}
and concatenate them to form $X_0 = [I_H; I_L] \in \mathbb{R}^{H \times W \times 2}$. The EAE processes $X_0$ through three identical blocks.As shown in Figure~\ref{eae}. Each block applies a $3 \times 3$ convolution, LayerNorm, GELU activation, and $2 \times 2$ max-pooling sequentially. Let $Z_k$ denote the output after the convolution, normalization, and activation at stage $k$. Then
\begin{align}
    Z_k &= \mathrm{GELU}\left( \mathrm{LayerNorm}\left( \mathrm{Conv}_{3\times3}(X_{k-1}) \right) \right), \\
    X_k &= \mathrm{MaxPool}_{2\times2}(Z_k),
\end{align}
for $k = 1, 2, 3$. 
To enhance contraband localization, we further process $X_3$ with location initializer to extract discriminative spatial features. We first flatten $X_3$ and apply a channel-wise linear head to generate pixel-wise attention weights. These weights are sorted in descending order, and the Top-$k$ corresponding feature embeddings are selected. The selected embeddings are then projected into the query space via a linear layer to initialize the output token of the Mask Decoder.

\subsection{Adaptive Point Generator}
In standard SAM, point prompts are randomly sampled within ground-truth masks during training. However, a single user-provided point often fails to capture fine-grained structures in X-ray contraband (e.g., a knife with visually distinct blade and handle due to density variations), leading to incomplete segmentation. To address this, we propose an \textbf{\textit{Adaptive Point Generator (APG)}} that transforms a single Naive Point \( p_0 \in \mathbb{R}^2 \) into two informative Finding Points \( \{p_1, p_2\} \).

Given input image \( I \) and \( p_0 \), APG first obtains an initial soft mask \( M_0 = \mathrm{SAM}(I; p_0) \). Let \( \mathcal{B}_0 = \mathrm{bbox}(M_0) \) denote its axis-aligned bounding box. To mitigate potential misalignment with the true object extent, we stochastically scale \( \mathcal{B}_0 \) by factors \( s_w, s_h \sim \mathcal{U}(0.9, 1.1) \), yielding an augmented region:
\begin{equation}
    \mathcal{B} = \mathrm{scale}(\mathcal{B}_0; s_w, s_h).
\end{equation}

Within \( \mathcal{B} \), we apply fast K-means clustering on foreground pixels of \( M_0 \) with \( K=2 \), obtaining centroids \( c_1, c_2 \in \mathbb{R}^2 \). If the clusters are well-separated (\( \|c_1 - c_2\|_2 > \tau \), with small threshold \( \tau \)), we sample one point per cluster and refine them to maximize spatial coverage:
\begin{equation}
    (p_1^\ast, p_2^\ast) = \arg\max_{\substack{p_1 \in \mathcal{C}_1 \\ p_2 \in \mathcal{C}_2}} \|p_1 - p_2\|_2,
\end{equation}
where \( \mathcal{C}_i = \{ x \in \mathcal{B} \mid \arg\min_{k} \|x - c_k\| = i-1 \} \) denotes the pixel set of cluster \( i \).

If only one cluster is detected (indicating appearance homogeneity), we fall back to uniform random sampling:
\begin{equation}
    p_1, p_2 \sim \mathrm{Uniform}(\mathcal{B}).
\end{equation}

The resulting prompt set \( \{p_1^\ast, p_2^\ast\} \) is fed into SAM’s prompt encoder, providing enhanced spatial guidance for heterogeneous contraband and significantly improving segmentation accuracy without additional user interaction.

\section{Experiments}
\label{sec:Experiments}

We conduct extensive experiments on the XSeg benchmark to rigorously evaluate the effectiveness of our proposed method. The evaluation framework encompasses implementation details, performance metrics, comparative analysis against state-of-the-art segmentation approaches, ablation studies, and qualitative segmentation visualizations. Our results demonstrate both the robustness of the XSeg dataset in large-scale security inspection tasks and the superior performance of APSAM in X-ray contraband segmentation, particularly in handling occlusions, multi-material compositions, and cluttered baggage scenarios.

\subsection{Implementation Details}

{\bf Experimental Setup:} Our experiments are implemented using PyTorch 1.8+ and conducted on a machine equipped with four NVIDIA RTX 3090 GPUs. To ensure fair comparison, all methods are trained exclusively on the XSeg training set and evaluated on the held-out XSeg test set. Our approach adopts a ViT-L/14~\cite{vit} backbone network. Given the average image resolution of 500×500 in XSeg, we standardize all inputs to 512×512 during training. The optimizer employs AdamW with an initial learning rate of $1e^{-5}$, a batch size of 16, and 12 training epochs, supplemented by a 100-step linear warmup strategy for stable convergence.

{\bf Evaluation Metrics:}  For performance evaluation, we adopt the widely recognized Dice coefficient and Intersection-over-Union (IoU) metrics, consistent with industrial inspection and security screening benchmarks. The Dice metric emphasizes overlap robustness against class imbalance, while IoU quantifies spatial consistency between predicted and ground-truth regions. Their combined use provides a comprehensive assessment of segmentation accuracy, particularly in challenging security scenarios involving occluded or cluttered contraband. All protocols were run three times independently using different random seeds, and the average metrics of the three experiments are ultimately reported.

\subsection{Comparison Experiments}

Given the current absence of segmentation algorithms specifically designed for X-ray contraband, we compared our approach against common CNN and Transformer-based segmentation algorithms adapted from other domains. Additionally, we conducted a brief comparative analysis involving frozen SAM~\cite{sam}, fine-tuned SAM, fine-tuned SAMUS~\cite{samus}, and our proposed method. Our APSAM demonstrates a significant advantage on the XSeg test set, with the detailed comparative results presented in Table~\ref{tab2}.

\begin{table*}[h]
\centering
\caption{Comparison of APSAM with other segmentation methods, including Frozen variants, where all parameters are kept frozen for inference only, and Finetune counterparts, which are trained using the identical adapter strategy as APSAM.}

\begin{tabular}{c|c|cccc}
\hline
Methods &Backbone  & mIoU(\%) & Dice(\%) & Trainable Params(M) & Pub'Year \\ \hline
PSPNet~\cite{pspnet}&ResNet101  &54.24  &70.34  &65.59  &CVPR'17  \\
DeepLabV3+~\cite{deeplab}&ResNet101  &57.29  &72.84  &60.21  & ECCV'18 \\
UperNet~\cite{upernet}&Swin-Tiny  &55.16  &71.10  &58.94  &ECCV'18  \\
Segformer~\cite{segformer}&MiT-b5&63.67  &77.80  &82.01  &NIPS'21  \\
Segmenter~\cite{segmenter}&ViT-B/16  &63.27  &77.50  &104.45  &ICCV'21  \\
Mask2former~\cite{mask2former} &Swin-L  &69.59  &81.44  &144.85  &CVPR'22  \\
Twins~\cite{twins}&PCPVT-L  &61.21  &75.94  &64.72  &NIPS'23  \\
MaskDINO~\cite{maskdino}&ResNet101  &65.22  &78.95  &62.90  &CVPR'23  \\
SegMAN~\cite{segman} &SegMANEncoder-b  &67.97  &77.09  &56.34  &CVPR'25  \\
SDANet~\cite{pidray} &ResNet101  &68.15  &80.46  &103.20  &ICCV'21  \\ \hline
SAM (Frozen)~\cite{sam} &ViT-L/14&53.82  &64.99  &0  &ICCV'23  \\
SAM (Finetune)~\cite{sam} &ViT-L/14  &67.87  &77.45  &10.06  &ICCV'23  \\
SAMUS (Finetune)~\cite{samus}&ViT-L/14  &68.56  &78.46  &43.21  &MICCAI'24  \\ \hline
\rowcolor{blue!10}
APSAM (Ours) &ViT-L/14  &\textbf{72.83}  &\textbf{82.31}  &11.91  & -  \\ \hline
\end{tabular}
\label{tab2}
\end{table*}

Compared to fine-tuning SAM alone, APSAM achieved improvements of \textbf{4.96\%} in mIoU and \textbf{4.86\%} in Dice. Furthermore, despite its trainable parameters being 31.3M fewer than SAMUS, a model with a larger parameter count and stronger performance, APSAM still demonstrated enhancements of \textbf{4.27\%} in mIoU and \textbf{3.85\%} in Dice.

\subsection{Ablation Studies}

We conducted several ablation studies to validate the reliability of our proposed Energy-Aware Encoder (EAE) and Adaptive Point Generator (APG), with the results detailed in Table~\ref{tabab1}. Without the EAE, mIoU and Dice scores were merely 71.90\% and 81.62\%, respectively. Similarly, excluding the APG resulted in mIoU and Dice scores of only 70.89\% and 79.50\%. However, when both EAE and APG were utilized concurrently, we observed a significant performance gain, with mIoU reaching \textbf{72.83\%} and Dice achieving \textbf{82.31\%}, representing an increase of \textbf{4.96\%} and \textbf{4.86\%} over the baseline. This robustly confirms the effectiveness of both our proposed EAE and APG components.

In addition, we conducted a simple ablation study on point prompts. Specifically, we compared three types of point prompting strategies: (1) directly using randomly selected point prompts provided by the user; (2) simulating the scenario where the user randomly provides two point prompts within the approximate region of the contraband; and (3) our proposed APG method. The experimental results are presented in Table~\ref{tabab2}. Under the condition of providing two point prompts, APG outperforms the random selection of two point prompts. These results demonstrate the superiority of APG. Moreover, APG is highly user-friendly, as it only requires a single user-provided point prompt and can adaptively determine the position of the second point prompt.

\begin{table}[h]
\centering
\caption{Ablation studies on our proposed EAE and APG demonstrate that optimal performance is achieved when both EAE and APG are utilized concurrently.}
\begin{tabular}{c|cc}
\hline
Methods & mIoU(\%) & Dice(\%)  \\ \hline
w/o EAE &71.90  &81.62    \\
w/o APG &70.89  &79.50    \\
w/o EAE \& APG &67.87  &77.45  \\ \hline
\rowcolor{blue!10}
w/ EAE \& APG &\textbf{72.83}  &\textbf{82.31} \\ \hline
\end{tabular}
\label{tabab1}
\end{table}

\begin{table}[h]
\centering
\caption{Ablation study of the APG, we evaluated the performance of randomly sampling a single point prompt, randomly sampling two point prompts, and our proposed APG. Our results clearly demonstrate that APG offers a significant advantage when utilizing two point prompts, outperforming simple random sampling.}
\begin{tabular}{c|cc}
\hline
Prompt Type & mIoU(\%) & Dice(\%)  \\ \hline
Random 1 point &67.87  & 77.45   \\
Random 2 Point &70.31  &80.18    \\ \hline
\rowcolor{blue!10}
APG (Ours) & \textbf{71.90}  &\textbf{81.62}   \\ \hline

\end{tabular}
\label{tabab2}
\end{table}

\subsection{Visualization of Segmentation}

\begin{figure}[htbp]
\centering
\includegraphics[width=\columnwidth]{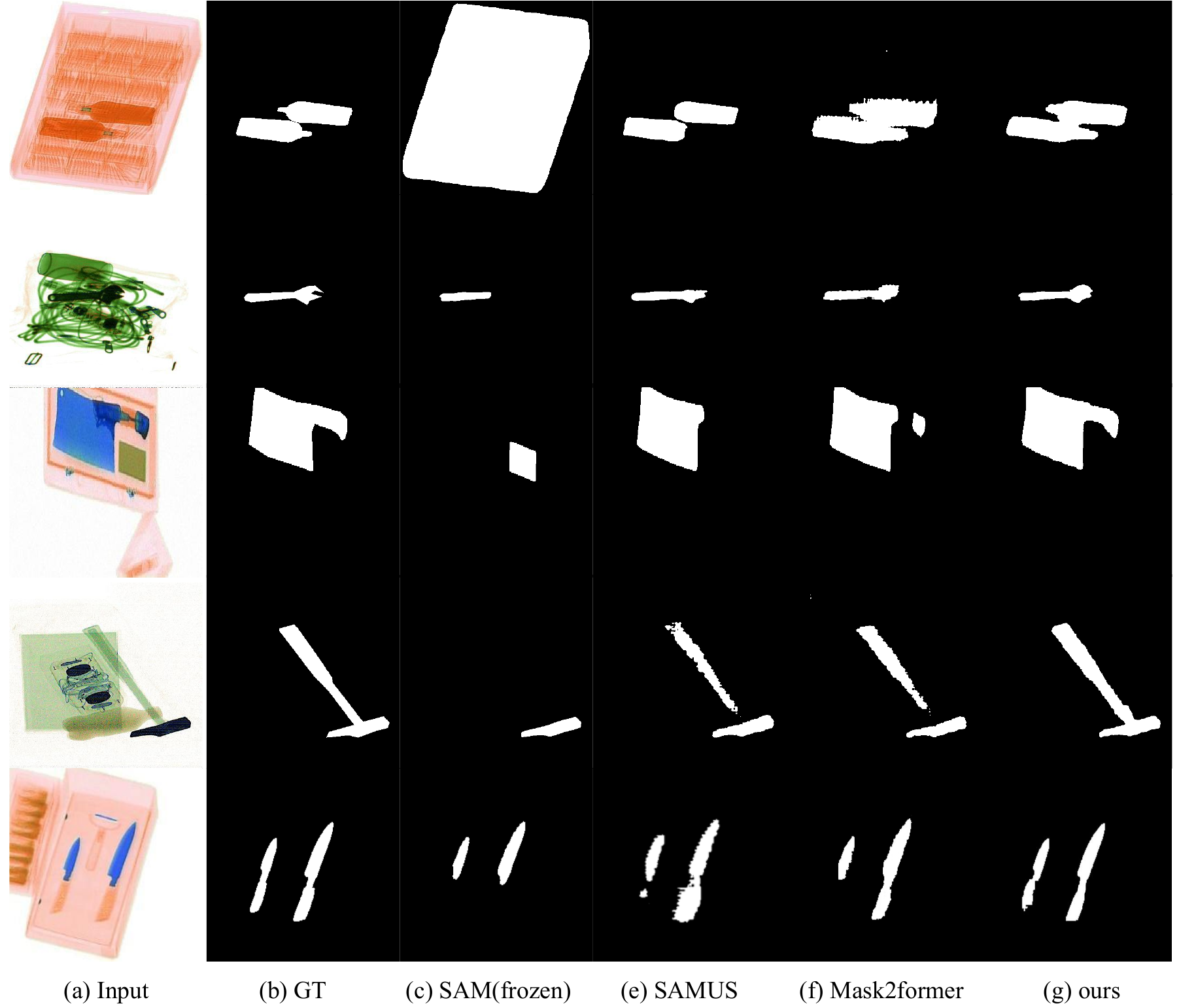}
\caption{Visualization of segmentation results. (g) shows the segmentation results of APSAM, which provides more complete and clearly visible segmentation boundaries than other schemes.}
\label{figviss}
\end{figure}

Figure~\ref{figviss} illustrates the segmentation performance of APSAM on the XSeg test set, primarily in comparison to other SAM-type segmentation methods. We first evaluated the segmentation capability of the original, frozen SAM. As shown in the first row of Figure~\ref{figviss}, the frozen SAM tends to segment regions with consistent color. When contraband and background objects have similar densities, the original frozen SAM only segments the background, proving ineffective for concealed items. Furthermore, as depicted in the second row of Figure~\ref{figviss}, if the user's initial point prompts are concentrated on a specific part of a prohibited item, SAM fails to segment the entire object. This limitation stems from its lack of prior knowledge about contraband imagery, resulting in segmentation outputs that lack holistic perception.

While SAMUS fine-tuned on XSeg shows preliminary awareness of contraband (Figure~\ref{figviss}, SAMUS column), its segmentation boundaries remain coarse and often fail to capture fine-grained structures. For example, in the third row, it struggles to distinguish the metal blade from the plastic handle of a kitchen knife, yielding fragmented and unsmooth masks. It also occasionally misses highly concealed objects, as seen in the fourth row.

In contrast, our APSAM significantly outperforms the previous state of the art, Mask2Former~\cite{mask2former}. APSAM produces continuous, smooth boundaries that closely align with true contraband edges and reliably detects even well-concealed threats, demonstrating superior structural understanding and robustness.

\subsection{Evaluation on Other Datasets}
\begin{table}[h]
\small
\centering
\caption{Evaluation results on PIDray and PIXray datasets. APSAM still maintains SOTA performance on the test sets of other datasets.}
\label{tabab3}
\resizebox{\columnwidth}{!}{
\begin{tabular}{l|cc|cc}
\hline
\multirow{2}{*}{Methods} & \multicolumn{2}{c|}{PIDray} & \multicolumn{2}{c}{PIXray} \\ \cline{2-5} 
 & mIoU (\%) & Dice (\%) & mIoU (\%) & Dice (\%) \\ \hline
DeepLabV3+~\cite{deeplab} & 64.62 & 73.34 & 73.17 & 86.36 \\
Mask2former~\cite{mask2former} & 69.04 & 78.18 & 78.36 & 86.12 \\
SegMAN~\cite{segman} & 66.51 & 75.24 & 78.37 & 87.87 \\
SAM~\cite{sam} & 64.54 & 75.10 & 66.20 & 77.82 \\
SAMUS~\cite{samus} & 67.01 & 76.85 & 80.63 & 88.26 \\ \hline
\rowcolor{blue!10}
APSAM (Ours) & \textbf{71.23} & \textbf{80.55} & \textbf{83.61} & \textbf{90.36} \\ \hline
\end{tabular}
}
\end{table}

Finally, we evaluate APSAM on the PIDray~\cite{pidray} and PIXray~\cite{pixray} to demonstrate its cross-domain generalization capability, and the results are shown in Table~\ref{tabab3}. Compared with SAMUS~\cite{samus}, APSAM achieved mIoU gains of \textbf{4.22\% }and \textbf{3.70\%} respectively on PIDray and PIXray, and Dice increased by \textbf{2.98\%} and \textbf{2.10\%} respectively. The experimental results show that even if APSAM is trained with only a partially filtered training set, it still manages to have a significant advantage over other datasets, providing a viable migration baseline for X-ray contraband segmentation.

\section{Conclusion}
This paper presents the largest and most diverse X-ray contraband segmentation dataset to date, XSeg. Considering the limitations of SAM in the field of contraband segmentation, we introduce APSAM,a specialized mask annotation model for contrabands,which integrates two modules: the Energy-aware encoder (EAE) and the Adaptive Point Generator (APG). EAE significantly enhances the model's sensitivity to stacked items, and APG ensures the effectiveness of prompt points. The experimental results show that APSAM outperforms the state-of-the-art models in the X-ray contraband segmentation task and has achieved leading results on the comprehensive XSeg benchmark. 
\section{Acknowledgements}
This work was supported in part by the Basic Scientific Research Operating Expenses of Xi'an Jiaotong University under Grants XXJ042025001, the Science and Technology Project of Guangzhou under Grants 202103010003, the Science and Technology Project in key areas of Foshan under Grants 2020001006285, the Xijiang Innovation Team Project under Grants XJCXTD3-2019-04B.

{
    \small
    \bibliographystyle{ieeenat_fullname}
    \bibliography{main}
}

\clearpage
\setcounter{page}{1}
\maketitlesupplementary

\section{Dual-energy X-ray Imaging}
X-ray images are primarily obtained by measuring the transmitted intensity \(I\) of an X-ray beam with energy \(E\) after it passes through a material. According to the Beer Lambert law~\cite{beer}, the relationship between the incident intensity \(I_0\), the transmitted intensity \(I\), the attenuation coefficient \(\mu\), the atomic number \(Z\), the beam energy \(E\), and the material thickness \(d\) is given by:

\begin{equation}
    I = I_0 \cdot e^{-\mu(E, Z) \cdot d}
    \label{eq:beer_lambert}
\end{equation}

In X-ray security imaging, two images are typically acquired: one with low-energy X-rays (\(E_l\)) and another with high-energy X-rays (\(E_h\)). X-rays of different energies have distinct penetration abilities. Specifically, below 200 keV, the attenuation coefficient \(\mu\) is dominated by two mechanisms:

\begin{equation}
    \mu(E, Z) = \mu_{\text{p}}(E, Z) + \mu_{\text{c}}(E, Z)
    \label{eq:total_attenuation}
\end{equation}

\begin{itemize}
    \item The photoelectric effect at low energies (\(E_l\)), where the attenuation coefficient is given by:
    \begin{equation}
        \mu_{\text{p}}(E, Z) \propto \frac{Z^3}{E^3}
        \label{eq:photoelectric}
    \end{equation}
    
    \item Compton scattering at high energies (\(E_h\)), where the attenuation coefficient is given by:
    \begin{equation}
        \mu_{\text{c}}(E, Z) \propto \frac{1}{E}
        \label{eq:compton}
    \end{equation}
\end{itemize}

\begin{figure}[h!]
    \centering
    \includegraphics[width=\linewidth]{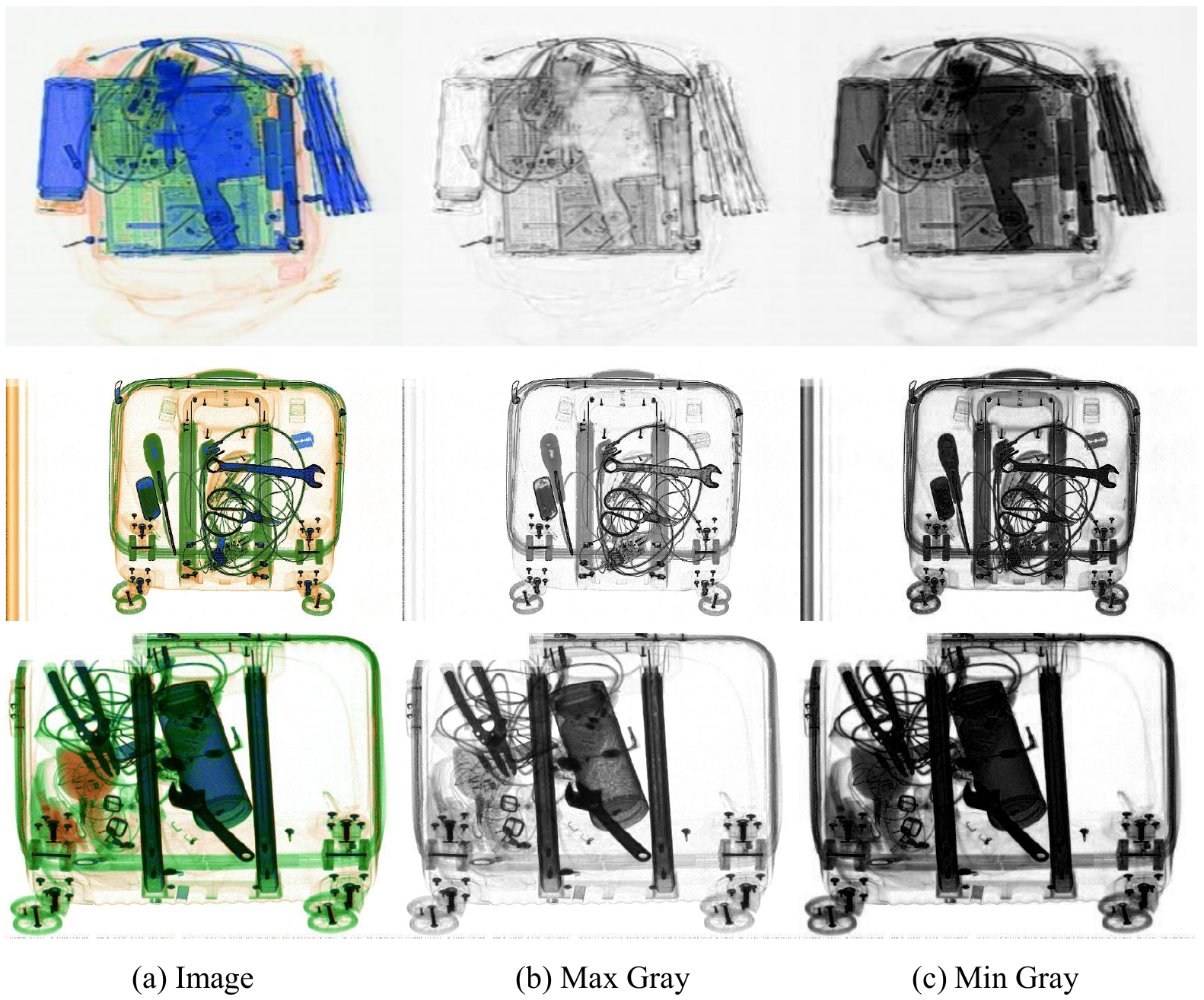}
    \caption{Display of the maximum and minimum grayscale images of contraband. The first column shows the contraband, the second column shows the corresponding maximum grayscale image, and the third column shows the corresponding minimum grayscale image.}
    \label{fig:dual}
\end{figure}

Both energy responses are measured for the same object, so the thickness \(d\) remains constant. For \(E_l\), the attenuation coefficient \(\mu\) is larger, resulting in a lower transmitted intensity \(I\), which corresponds to the minimum grayscale value. In contrast, for \(E_h\), the attenuation coefficient \(\mu\) is smaller, leading to a higher transmitted intensity \(I\), which corresponds to the maximum grayscale value \(G\) defined as shown in Equation~\ref{eq:max_grayscale}:
\begin{equation}
\begin{split}
    G_{\text{max}} &= \max(G_H, G_L) = G_H, \\
    G_{\text{min}} &= \min(G_H, G_L) = G_L
\end{split}
\label{eq:max_grayscale}
\end{equation}
The maximum and minimum grayscale images are illustrated in Figure~\ref{fig:dual}.

From the example in Figure~\ref{fig:dual}, it can be noticed that the maximum grayscale map tends to provide the overall location of the contraband in the contraband image, while the minimum grayscale map visually makes the localized location of the contraband affected by the stacking situation more prominent.

\section{Xseg Semantic Segmentation Labels}

\begin{figure*}[h!]
    \centering
    \includegraphics[width=\linewidth]{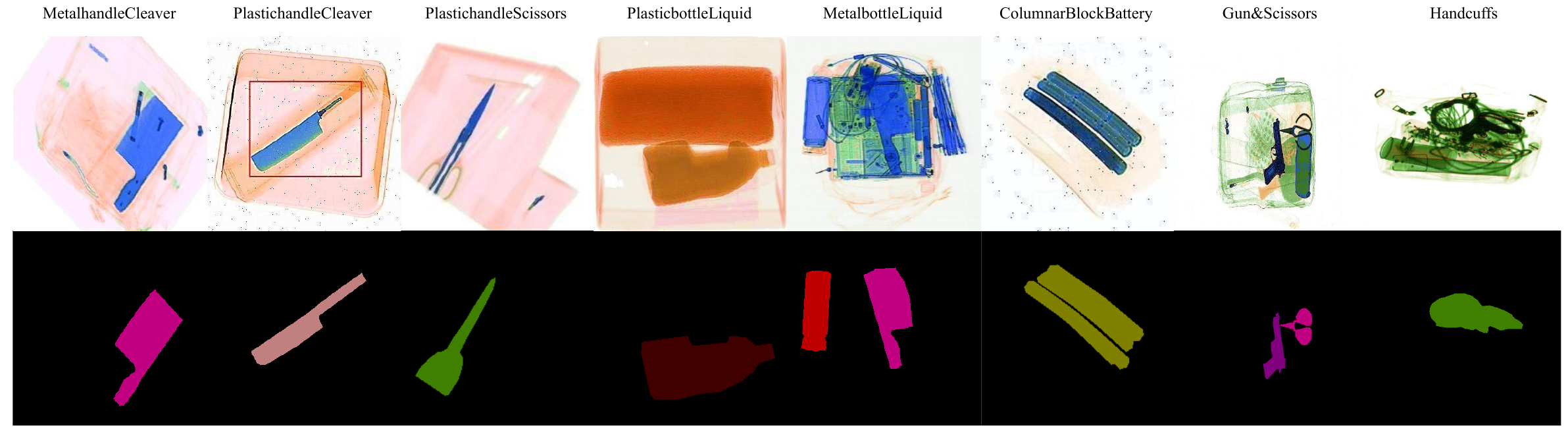}
    \caption{Hierarchical structure of fine-grained semantic annotations in the Xseg dataset.}
    \label{fig:finegrained_flowchart}
\end{figure*}

\begin{table*}[h!]
\centering
\caption{Fine-grained categories grouped by semantic superclasses.}
\label{tab:finegrained_categories}
\begin{tabular}{@{}p{3.5cm}p{12cm}@{}}
\toprule
\textbf{Superclass} & \textbf{Fine-grained Categories} \\
\midrule
Batteries & ColumnarBlockBattery, MotorBattery \\
Suspicious Liquid & PlasticbottleLiquid, GlassbottleLiquid, MetalbottleLiquid \\
Fruitknife & MetalhandleFruitknife, PlastichandleFruitknife \\
Cleaver & MetalhandleCleaver, PlastichandleCleaver \\
Scissors & MetalhandleScissors, PlastichandleScissors, Scissors \\
Batons & ExpandableBatons, Baton \\
Electronic Devices & Powerbank, Mobilephone, Laptop \\
Tools & Hammer, Pliers, Wrench, Screwdriver \\
Firearms & Gun, Bullet \\
Others & Pressure, Handcuffs, Lighter, Fireworks, Dart, Razorblade, Sawblade \\
\bottomrule
\end{tabular}
\end{table*}

In addition to traditional binary labels, we further provide more discriminative semantic segmentation labels for X-ray images to support more complex downstream vision tasks. On top of that, we introduce a fine-grained semantic annotation scheme, where certain key categories are further divided into multiple semantically consistent subcategories to more accurately describe their internal structures and functional differences. Such fine-grained labeling plays a crucial role in enhancing the model's ability to recognize objects in highly overlapping regions and under complex occlusion scenarios.

As illustrated in Table~\ref{tab:finegrained_categories}, we present the hierarchical design of our fine-grained categorization system, which clearly demonstrates the subdivision logic from coarse categories to specific subcategories. For example, we divide suspicious liquids into plasticbottle liquids, glassbottle liquids, and metalbottle liquids based on their color characteristics that reflect different material densities. For categories such as fruit knives, scissors, and cleaver, we further classify them into metal-handle and plastic-handle types according to the handle material. Similarly, batteries are grouped into columnarblock batteries and motor-attached batteries based on their appearance. This structured annotation not only enhances the expressive power of the labels but also provides clearer learning targets for the model.

To validate the effectiveness of the fine-grained annotations, we conduct a series of semantic segmentation and multi-class classification experiments based on this labeling system. We further compare the performance of models trained with fine-grained labels against those trained with coarse-grained labels, in order to assess the impact of fine-grained information on improving model discriminative ability.
\end{document}